\documentclass[lettersize,journal]{IEEEtran}
\usepackage{amsmath,amsfonts}
\usepackage{algorithmic}
\usepackage{algorithm}
\usepackage{array}
\usepackage[caption=false,font=normalsize,labelfont=sf,textfont=sf]{subfig}
\usepackage{textcomp}
\usepackage{stfloats}
\usepackage{url}
\usepackage{verbatim}
\usepackage{graphicx}
\usepackage{float}
\usepackage{todonotes}
\usepackage{multirow}
\usepackage{makecell}
\usepackage{color,array}
\usepackage{cite}
\usepackage[colorlinks,urlcolor=blue,linkcolor=blue,citecolor=blue]{hyperref}
\usepackage{enumerate}

\begin{document}

\title{LLMs as \textit{Repositories} of Factual Knowledge: \\ Limitations and Solutions} 

\author{Seyed Mahed Mousavi, Simone Alghisi, and Giuseppe Riccardi, \IEEEmembership{Fellow, IEEE}
\thanks{Seyed Mahed Mousavi, Simone Alghisi, and Giuseppe Riccardi are with Signals and Interactive Systems Lab, Department of Information Engineering and Computer Science, University of Trento, Italy. We acknowledge the support of the MUR PNRR project FAIR - Future AI Research (PE00000013) funded by the NextGenerationEU.}
}

\markboth{IEEE/ACM TRANSACTIONS ON AUDIO, SPEECH, AND LANGUAGE PROCESSING,~Vol.~xx, No.~x}%
{Shell \MakeLowercase{\textit{et al.}}: A Sample Article Using IEEEtran.cls for IEEE Journals}


\maketitle

\begin{abstract}
LLMs' sources of knowledge are data snapshots containing factual information about entities collected at different timestamps and from different media types (e.g. wikis, social media, etc.). Such unstructured knowledge is subject to change due to updates through time from past to present. Equally important are the inconsistencies and inaccuracies occurring in different information sources. Consequently, the model's knowledge about an entity may be perturbed while training over the sequence of snapshots or at inference time, resulting in inconsistent and inaccurate model performance. In this work, we study the appropriateness of Large Language Models (LLMs) as repositories of factual knowledge. We consider twenty-four state-of-the-art LLMs that are either closed-, partially (weights), or fully (weight and training data) open-source. We evaluate their reliability in responding to time-sensitive factual questions in terms of accuracy and consistency when prompts are perturbed. We further evaluate the effectiveness of state-of-the-art methods to improve LLMs' accuracy and consistency. We then propose “ENtity-Aware Fine-tuning” (ENAF), a soft neurosymbolic approach aimed at providing structured representation of entities during fine-tuning to reduce inconsistencies and improve response stability under prompt variations. 
\end{abstract}


\section{Introduction}
\IEEEPARstart{L}{arge} Language Models (LLMs) are transforming from research prototypes to a common go-to for querying a vast variety of factual knowledge. Such models are being compared to traditional knowledge repositories like knowledge bases \cite{cohen-etal-2023-crawling}, knowledge graphs \cite{sun2023head}, and search engines \cite{pinter-elhadad-2023-emptying}. LLMs' sources of knowledge are multiple data snapshots \(T_i\) collected at different timestamps \(i \in \{1, \dots, n\}\) \cite{dhingra-etal-2022-time}. Each snapshot \(T_i\) (e.g., \textit{Wikipedia dump 2017, Reddit dump 2018, etc.}) contains factual information about a set of entities \( E = \{ e_1, \dots, e_j, \dots, e_n \} \) (e.g., \textit{the football player Cristiano Ronaldo}) drawn from different sources. Each fact can be represented as a triplet \( F_i = (\text{\textit{subject}}, \text{\textit{property}}, \text{\textit{attribute}}) \) (e.g. \( (\text{Cristiano Ronaldo}, \text{plays at}, \text{Juventus}) \)).

Despite their capabilities, LLMs face fundamental challenges as reliable knowledge sources. The importance of reliability for a knowledge base lies in its role as a trustworthy source of information. Two important requirements for a reliable knowledge base are accuracy and consistency \cite{DIGNUM1992293}. Regarding \textbf{accuracy}, factual knowledge is inherently dynamic, continually evolving as new information is introduced and older data becomes outdated. Thus, a reliable knowledge repository must maintain the validity of its information, ensuring it remains accurate and up-to-date. Regarding \textbf{consistency}, a knowledge base must integrate new information without introducing contradictions, preserving the integrity and the logical coherence of the knowledge base. A knowledge base failing to fulfill these requirements risks misleading the users, leading to incorrect conclusions and decisions.

In the context of LLMs, such models are static artifacts, trained on fixed data snapshots. This makes the LLMs prone to generating outdated responses, quickly rendering them unreliable as sources of time-sensitive information. Furthermore, the data snapshots used for training may include different lexicalizations of \(e_j\) (e.g., \textit{Cris R., CR7, Ronaldo, Cristiano, etc.}) and frequently overlap \cite{soldaini2024dolma}. Thus, the knowledge about the entity \(e_j\) may be fragmented during training across snapshots \(T_1, T_2, \dots, T_n\) either due to real-world updates or inconsistencies between different information sources. Additionally, providing the most up-to-date information at inference time can further perturb this knowledge by introducing external information that may conflict with the model’s internal knowledge \cite{wu2024faithful}. Consequently, the knowledge of the model about entity \(e_j\)  may become crumbled, resulting in inconsistent responses when queried about the entity.

In this work, we study the reliability of LLMs as repositories of factual knowledge. Our study aims to identify the limitations and shortfallings of such models and experiment with novel approaches to improve their performance. We frame our investigations around two central research questions:

\textbf{RQ1 (Assessment): How Reliable State-of-the-Art LLMs are as Repositories of Time-Sensitive Facts?} 
LLMs are typically evaluated using static benchmarks. These benchmarks can quickly become outdated due to the evolving nature of knowledge, making them less effective for long-term evaluations. Besides, static benchmarks are prone to contamination \cite{balloccu-etal-2024-leak}, where benchmark data leaks into the training sets of future models, potentially biasing the results. To address these issues, we introduce \textbf{DyKnow}, a dynamic benchmarking framework that continuously updates data points using real-time information from Wikidata. Unlike static benchmarks, dynamic benchmarks like DyKnow provide ongoing assessments of LLMs’ knowledge, addressing the problems of outdatedness and data contamination. 

Via DyKnow, we assess the accuracy and consistency of twenty-four state-of-the-art LLMs across a diverse set of time-sensitive facts. We assess the \textbf{accuracy} of the models by evaluating model responses against the most current and comprehensive list of values retrieved from Wikidata at the evaluation time. We further examine the \textbf{consistency} of LLMs by evaluating their responses across multiple perturbations of the question prompts - an approach known as \textit{prompt agreement}, as a metric for measuring input-bound uncertainty \cite{leidinger-etal-2023-language, portillo-wightman-etal-2023-strength}. We assess output consistency under two types of perturbations: a) \textit{subject perturbations}, where different lexicalizations of the same entity \(e_j\) are used, comparing \(p(\text{attribute} | e_j, \text{property})\) with \(p(\text{attribute} | e_j^*, \text{property})\); and b) \textit{property perturbations}, which involve querying the model with different lexicalizations of the property to compare \(p(\text{attribute} | e_j, \text{property})\) to \(p(\text{attribute} | e_j, \text{property}^*)\). 

\textbf{RQ2 (Improvement): Can We Enhance the Reliability of LLMs as Time-Sensitive Knowledge Repositories?} Achieving reliability for the LLMs requires a systematic approach to identifying outdated/inconsistent information, and applying necessary changes. In this research question, we aim to identify the most effective methods for improving LLMs' reliability as dynamic, up-to-date knowledge repositories. We focus on five LLMs with identified outdated knowledge and inconsistent performance in RQ1 (Assessment). We experiment with different approaches to improve the accuracy and consistency of these LLMs as knowledge repositories. 

Firstly, we evaluate the effectiveness of state-of-the-art approaches that aim to improve the model performance without retraining/fine-tuning the model. We experiment with four knowledge editing mechanisms to update specific facts within the model: two approaches that modify LLM parameters, ROME \cite{meng2022locating} and MEMIT \cite{meng2022mass}, as well as two methods that preserve the original parameters, SERAC \cite{pmlr-v162-mitchell22a} and IKE \cite{Zheng2023CanWE}. We compare these methods with Retrieval-Augmented Generation (RAG) \cite{gao2023retrieval}, a technique that aims to improve the model performance by retrieving relevant external information during inference, potentially bypassing outdated internal knowledge. 

Next, we introduce \textit{ENtity-Aware Fine-tuning (ENAF)} a soft neurosymbolic approach designed to introduce structured representations of entities into the model via data annotation. Traditional (pre-)training and fine-tuning techniques generally lack structured entity representations, often leading to fragmented knowledge within the model, where different lexical variations of the same entity may not be consistently linked. We aim to address this issue by promoting a unified neurosymbolic representation of each entity, enabling the model to map different perturbations or lexical variations back to a single symbolic reference. Through the structured representations, ENAF aims to help the model achieve a more integrated understanding of entities, thereby reducing inconsistencies across prompt variations. We experiment with various structured representations such as Named Entity tags and unique entity ID tags. We investigate the impact of structured representations of entities on the model’s internal knowledge representation, compared to the unstructured text typically used to (pre-)train LLMs. 

By addressing these research questions, we aim to advance the understanding of LLMs' skill as knowledge {\em holder}, {\em retriever} and {\em manager}. Lastly, we propose neurosymbolic frameworks for future advancements in dynamic knowledge evaluation and management of LLMs. While our analysis centers on text-based LLMs, the examined issues, particularly consistency under input variation and factual reliability, are equally critical in speech-driven and multimodal settings. In spoken dialogue and ASR–LLM pipelines, utterance variability makes robustness to perturbations essential. Besides, dynamic evaluation of factual reliability benefits voice-based LLMs that require up-to-date world knowledge. Therefore, we believe the findings of this work naturally extend to speech-oriented and multimodal LLM architectures\footnote{This work builds upon our previous conference paper \cite{dyknow}, where we introduced DyKnow as a dynamic benchmarking for assessing the time-sensitive factual knowledge in LLMs, focusing on property perturbations, and evaluated the effectiveness of knowledge editing algorithms on four models. The current journal article provides in-depth and extended analyses, including consistency evaluations upon subject perturbations, comparative performance assessments between knowledge editing methods and Retrieval-Augmented Generation (RAG), and proposes a novel neurosymbolic approach ENtity-Aware Fine-tuning (ENAF), designed to enhance model consistency. All accompanying materials will also be made available to support further use and extension of this work at \url{https://github.com/sislab-unitn/DyKnow}.}.

\section{Literature Review}

\textbf{LLMs.} Current LLMs face significant challenges as knowledge repositories, particularly in areas such as editing, logical consistency, reasoning, and interoperability \cite{pinter-elhadad-2023-emptying}. Although knowledge editing methods have been proposed to update LLMs' knowledge, most focus on synthetic counterfactual target datasets and often do not scale well to real-world scenarios \cite{zhang-etal-2023-large}. Studies show that existing knowledge editing techniques are prone to several issues, including catastrophic forgetting \cite{ratcliff1990connectionist}, restrictions on the number of permissible edits \cite{mitchell2021fast}, ripple effect failures \cite{cohen2023evaluating}, and a lack of robustness \cite{brown2023robustness, hase2023does}. On the other hand, research on aligning LLMs with real-world knowledge reveals several shortcomings, including unrealistic evaluation scenarios, reliance on synthetic datasets, inadequate quantitative analysis, and more importantly a gap in detecting outdated knowledge in LLMs \cite{zhang-etal-2023-large}. A comprehensive review of model editing across both computer vision and NLP has been provided in \cite{mazzia2023survey}.

\textbf{Benchmarks.} The benchmarks for evaluating LLMs' knowledge and temporal reasoning often refer to a specific point in the past using explicit time-specifiers \cite{chen2021dataset, gupta-etal-2023-temptabqa}, or in more complex scenarios, multiple temporal factors \cite{wei-etal-2023-menatqa}. An evaluation framework assessing models on memorization, understanding, application, and creation of knowledge was proposed in \cite{yu2023kola}. Challenges in building dynamic factual benchmarks were discussed in \cite{yin-etal-2023-alcuna}, where the authors suggested generating artificial knowledge by randomly altering entities or relations within the same ontological class. While these studies focused on static benchmarks, a more dynamic benchmark, RealTime QA, which features 30 weekly questions and answers for LLM evaluation, was introduced in \cite{kasai2022realtime}. Another approach \cite{jang-etal-2022-temporalwiki} tracks changes in knowledge by comparing consecutive Wikipedia snapshots and re-training models based on the identified differences.

\section{Assessment of LLMs' Reliability (RQ1)}

There is a notable gap in detecting outdated knowledge in LLMs, as identified in prior research \cite{zhang-etal-2023-large}. A benchmark for evaluating time-sensitive knowledge in LLMs must be both model-agnostic and long-lasting, ensuring its relevance as models evolve. However, LLMs are typically evaluated using static benchmarks which are unable to capture the dynamic nature of real-world knowledge, posing a risk of becoming outdated, and contaminating the training data of future models. To address these limitations, dynamic benchmarking has been proposed, where data points are continuously updated to reflect real-time information. However, despite extensive work on static benchmarks, theoretical and empirical research on dynamic benchmarking remains limited \cite{shirali2023a}. Consecutively, constructing reliable dynamic benchmarks remains challenging and expensive \cite{yin-etal-2023-alcuna}. 

\subsection{\textbf{DyKnow}: \textbf{Dy}namic \textbf{Know}ledge Benchmark}

We present DyKnow, a dynamic benchmarking framework that updates its data points with real-time information from Wikidata. In contrast to static benchmarks, DyKnow enables ongoing evaluation of LLMs, effectively reducing the risks of outdated knowledge and data contamination.

In Wikidata \cite{vrandevcic2014wikidata}, factual knowledge is structured through \textit{properties} that link \textit{subject} nodes to corresponding \textit{attribute} values. The attribute values in Wikidata are further enriched with \textit{qualifiers} that provide additional context, including start and end dates that define the temporal validity of the attribute. Wikidata also maintains historical records of previous attribute values, along with their associated start and end dates, indicating the periods during which those values were valid. For example, the fact "\textit{Cristiano Ronaldo's current football club}" is represented by the property {\fontfamily{cmtt}\selectfont{"member of sports team"}} connecting the subject {\fontfamily{cmtt}\selectfont{"Cristiano Ronaldo"}} to the current attribute value, "{\fontfamily{cmtt}\selectfont{Al-Nassr\textsubscript{2023-Now}}}," accompanied by {\fontfamily{cmtt}\selectfont{"Manchester United F.C. \textsubscript{2021-2022}", "Juventus FC \textsubscript{2018-2021}", etc}}. This rich temporal data extracted from Wikidata allows us to effectively evaluate LLMs with real-time and historically accurate information. 

DyKnow dynamically evaluates the models' outputs against attribute values retrieved from the Wikidata knowledge base at the time of evaluation, instead of relying on static ground truths. The model's knowledge for each fact is assessed as \textit{correct} when it outputs the most current value from the list. For \textit{incorrect} outputs, we further classify them as \textit{outdated} when the model provides a response that was once correct but is now outdated, and \textit{irrelevant} when the model generates a value not found in the Wikidata list, often due to hallucination or errors in the training data. 

\textbf{Facts.} We evaluate the reliability of the models regarding time-sensitive facts in different categories. Each fact is presented as triplet of \( F_i = (\text{\textit{subject}}, \text{\textit{property}}, \text{\textit{attribute}})\). Regarding subject selection in this version of DyKnow, we focus on entities that are likely to appear frequently in the training data of most LLMs. Prior work shows that an LLM’s ability to recall factual information depends strongly on the frequency with which an entity is seen during pre-training \cite{pinter-elhadad-2023-emptying, mallen-etal-2023-trust}. Concentrating on high-frequency entities, therefore, enables reliable and interpretable evaluation of up-to-dateness, as it reduces ambiguity about whether model errors arise from outdated knowledge, inconsistencies in parametric memory, or the simple absence of a training signal. We acknowledge that this excludes many low-frequency or long-tail entities, where behavior may differ substantially. However, DyKnow is fully extensible, and all resources have been released to support future expansions targeting rare or domain-specific entities. To ensure broad coverage, we select entities based on their prominence in various domains: the top 50 countries by Gross Domestic Product (GDP) in 2023, the top 30 athletes (10 soccer players, 10 basketball players, and 10 Formula 1 drivers), and 25 leading public and private organizations (including the top 20 companies by revenue and the top 5 organizations by influence). Regarding properties, we query the models regarding the "\textit{head of state}" (e.g., president, king) and the "\textit{head of government}" (e.g., prime minister, premier) for each country. Meanwhile, for each athlete, the query focuses on their \textit{current sports team}, while for organizations, the query concerns the \textit{directorial role} (e.g., CEO, chairperson). After manually filtering out subjects with incomplete property or attribute data in Wikidata, the final set of time-sensitive facts used to benchmark the LLMs includes 78 facts covering 47 countries, 28 facts about athletes, and 24 facts concerning 23 organizations. Therefore, we evaluate the models regarding 130 time-sensitive facts about frequent subject entities in different categories (human subjects, organization subjects, and country subjects). The complete list of subject entities and properties, along with the time-sensitive questions, is presented in our prior work (Appendix Table 3) \cite{dyknow}.
\begin{table}[t!]
\caption{The Accuracy of various Large Language Models (LLMs) on 130 time-sensitive facts in DyKnow benchmark.}
\label{table:full_currency}
\centering
\begin{tabular*}{17.85pc}{p{90pt} |p{25pt}<{\raggedleft\hangindent5pt}
                                p{25pt}<{\raggedleft\hangindent5pt}
                                p{25pt}<{\raggedleft\hangindent5pt}}
\hline
\multirow{2}{*}{\textbf{(Year) Model}} & \multirow{2}{*}{\textbf{C}{orrect}} & \multicolumn{2}{c}{\textbf{Incorrect}} \\
\cline{3-4}
 & & \textbf{O}{utdated} & \textbf{I}{rrelevant} \\
\hline\\ [-8pt]
{\small(2019)} GPT-2 & 26\% & 42\% & 32\%\\
{\small(2020)} GPT-3 & 42\% & 47\% & 12\%\\
{\small(2020)} T5 & 11\% & 21\% & 68\%\\
{\small(2021)} GPT-J & 41\% & 46\% & 13\%\\
{\small(2022)} Bloom & 35\% & 49\% & 16\%\\
{\small(2022)} Flan-T5 & 18\% & 39\% & 43\%\\
{\small(2023)} Llama-2 & 51\% & 42\% & 7\%\\
{\small(2023)} Falcon & 42\% & 47\% & 11\%\\
{\small(2023)} Mistral & 53\% & 39\% & 8\%\\
{\small(2023)} Mixtral & 48\% & 42\% & 10\%\\
{\small(2024)} OLMo \small{1B} & 37\% & 40\% & 23\%\\
{\small(2024)} OLMo \small{7B} & 35\% & 36\% & 29\%\\
{\small(2024)} Llama-3 & 57\% & 36\% & 7\%\\
{\small(2024)} OpenELM \small{270M} & 12\% & 28\% & 61\%\\
{\small(2024)} OpenELM \small{1.1B} & 35\% & 47\% & 18\%\\
{\small(2024)} OpenELM \small{3B} & 42\% & 42\% & 16\%\\
\hline\\ [-8pt]
{\small(2022)} ChatGPT & 57\% & 35\% & 8\%\\
{\small(2023)} GPT-4 & 80\% & 13\% & 7\%\\
{\small(2023)} Llama-2$_{C.}$ & 51\% & 37\% & 12\%\\
{\small(2023)} Falcon$_{I.}$ & 44\% & 41\% & 15\%\\
{\small(2023)} Vicuna & 52\% & 33\% & 15\%\\
{\small(2023)} Mistral$_{I.}$ & 52\% & 32\% & 16\%\\
{\small(2023)} Mixtral$_{I.}$ & 62\% & 29\% & 9\%\\
{\small(2024)} Llama-3$_{I.}$ & 76\% & 14\% & 10\%\\
\hline
\multicolumn{4}{p{17pc}}{Models, accompanied by the corresponding release year, are evaluated via \textit{Upper Bound}. LLMs below the line were prompted with an additional prefix "Answer with the name only". Subscripts ${I.}$ and ${C.}$ stand for \textit{Instruct} and \textit{Chat}, respectively.}\\
\end{tabular*}
\end{table}

\textbf{Models.} Via DyKnow, we evaluate the accuracy of the following 24 LLMs: GPT-2 XL \cite{radford2019language}, GPT-3\footnote{We used \href{https://platform.openai.com/docs/models/gpt-base}{'davinci-002'} model for GPT-3.} \cite{brown2020language}, T5 (3B) \cite{raffel2020exploring}, GPT-J (6B) \cite{gpt-j}, ChatGPT (GPT-3.5)\footnote{We used \href{https://platform.openai.com/docs/models/gpt-3-5}{'gpt-3.5-turbo-1106'} for ChatGPT (GPT-3.5).}, Bloom (7B) \cite{workshop2022bloom}, Flan-T5 XL \cite{chung2022scaling}, GPT-4\footnote{We used \href{https://platform.openai.com/docs/models/gpt-4-and-gpt-4-turbo}{'gpt-4-1106-preview'} for GPT-4.}, Llama-2 (7B) \& Llama-2 Chat (7B) \cite{touvron2023llama}, Falcon (7B) \& Falcon Instruct (7B) \cite{almazrouei2023falcon}, Vicuna v1.5 (7B) \cite{vicuna2023}, Mistral v0.1 (7B) \& Mistral Instruct v0.1 (7B) \cite{jiang2023mistral}, Mixtral 8x7B v0.1 \& Mixtral 8x7B Instruct v0.1 \cite{jiang2024mixtral}, OLMo (1B \& 7B) \cite{groeneveld2024olmo}, Llama-3 (8B) and Llama-3 Instruct (8B)\footnote{The instance used for LLaMA-3 is accessible by this \href{https://ai.meta.com/blog/meta-llama-3/}{Link}.}, OpenELM (270M \& 1.1B \& 3B)~\cite{mehta2024openelm}.

\label{promptstrategysection}
\textbf{Prompting Strategy.} We create a prompt template tailored for each fact and subject group, with placeholders for subject names and, for countries, official titles. Using GPT-4, we generate four rephrased versions of each prompt with slight lexical variations and have three human judges (researchers from our team) review and validate the generated prompts. After gathering feedback and conducting manual checks, three final question prompts are selected for each fact. The models are then queried using these three selected prompts for each time-sensitive fact. Unlike prior studies on LLM temporal reasoning \cite{chen2021dataset, wei-etal-2023-menatqa, gupta-etal-2023-temptabqa}, our prompts are framed in the present tense, exclude explicit time specifiers, and aim to elicit the \textit{currently correct} answer.  The prompt templates used for querying the models with time-sensitive facts across different subject categories are outlined in our prior work (Appendix Table 3) \cite{dyknow}.

\subsection{Evaluating LLMs' Accuracy} 

\textbf{Upper Bound.} We validate the model's outputs using an "\textit{Upper Bound}" strategy. If the model provides the correct (most current) answer in response to at least one of the three prompts, we consider it successful, demonstrating that the model holds up-to-date information for that particular fact. If the model fails to return a correct answer but produces an outdated response for at least one of the prompts, we classify the model’s knowledge of the fact as outdated. Meanwhile, irrelevant responses can occur for several reasons: a) the model may not have encountered the specific time-sensitive fact during (pre-)training or fine-tuning, b) hallucinations or the presence of conflicting/false data in the training set, or c) the fact may exist in the model but is inaccessible through the prompts used.

\textbf{Results.} Table~\ref{table:full_currency} shows the results of this evaluation, indicating a substantial portion of model outputs are either outdated or irrelevant. These findings reveal significant concerns about the accuracy of time-sensitive knowledge across these models, posing challenges for real-world applications if these models are relied upon. Even the best-performing models show non-negligible percentages of outdated and irrelevant answers. GPT-4 (2023) performs well, with a high proportion of correct responses, but still produces outdated or irrelevant answers in 20\% of cases. Similarly, more recent models like Llama-3 (2024), OLMo (2024), and OpenELM (2024) provide incorrect responses-either outdated or irrelevant-for over 40\% of the queries. As anticipated, older models such as GPT-2 (2019) and GPT-3 (2020) exhibit lower levels of up-to-date knowledge.

\subsection{Evaluating LLMs' Consistency}

\begin{figure}[t]
\centering
    \includegraphics[trim={0.2cm 0.6cm 0 0},clip,width=0.8\linewidth]{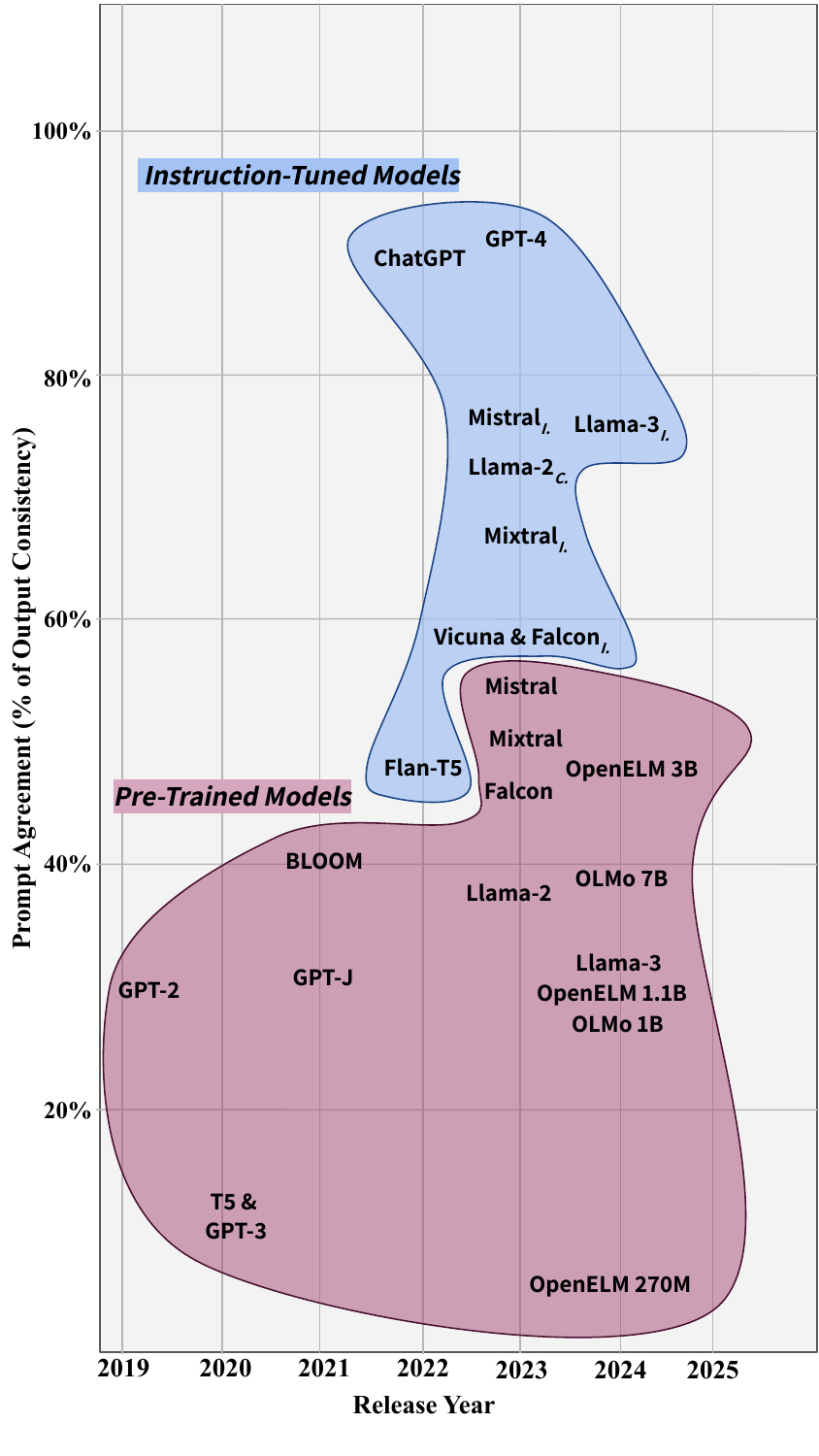}
    \caption{Output consistency across LLMs when prompted with \textit{subject perturbations}. The x-axis represents the release year of each model, and the y-axis shows the level of output consistency when prompted with varied lexicalizations of the same subject. Subscripts ${I.}$ and ${C.}$ stand for \textit{Instruct} and \textit{Chat}, respectively. Instruction-tuned models demonstrate a comparatively higher prompt agreement.}
    \label{subjectconsistencyfig}
\end{figure} 

We assess the output consistency of LLMs across multiple prompts as input-bound uncertainty of the model \cite{leidinger-etal-2023-language,portillo-wightman-etal-2023-strength}. Output consistency is evaluated under two conditions: a) \textit{subject perturbations}, where different lexicalizations of the same subject entity \(e_j\) are used, comparing \(p(\text{attribute} | e_j, \text{property})\) with \(p(\text{attribute} | e_j^*, \text{property})\); and b) \textit{property perturbations}, which involve alternative lexicalizations of the property to compare \(p(\text{attribute} | e_j, \text{property})\) with \(p(\text{attribute} | e_j, \text{property}^*)\).

\begin{figure}[t]
\centering
    \includegraphics[trim={0.2cm 0.6cm 0 0},clip,width=0.75\linewidth]{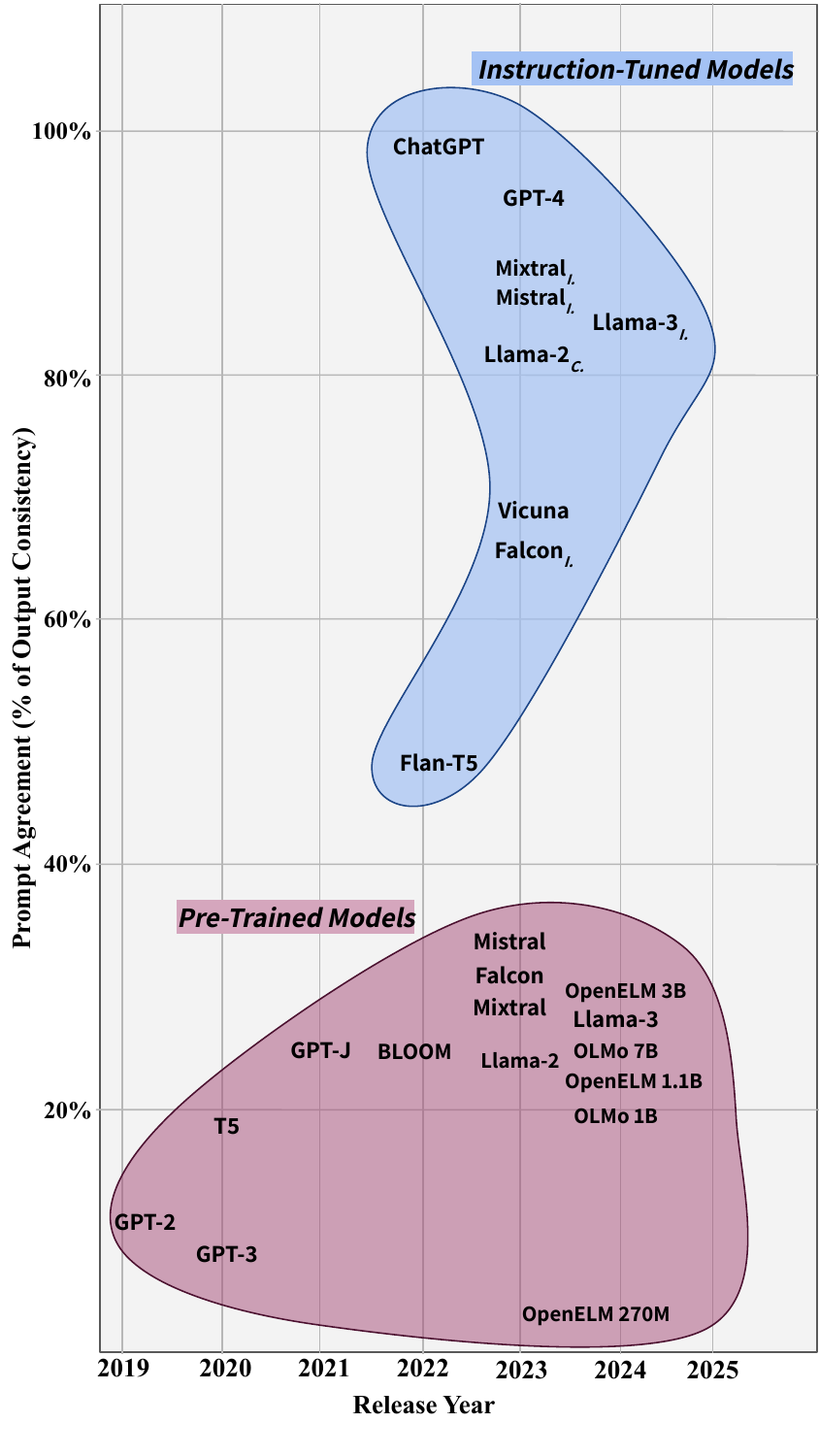}
    \caption{Output consistency across LLMs when prompted with \textit{property perturbations}. The x-axis represents the release year of each model, and the y-axis shows the level of output consistency when prompted with varied lexicalizations of the property. Subscripts ${I.}$ and ${C.}$ stand for \textit{Instruct} and \textit{Chat}, respectively. Instruction-tuned models demonstrate a comparatively higher prompt agreement.}
    \label{propertyconsistencyfig}
\end{figure} 

\textbf{Subject Perturbation.} We use GPT-4 to generate five variations of each subject name (e.g. aliases) as lexical perturbations. These variations are then reviewed and validated by three human judges (researchers from our team) to ensure their appropriateness and validity. After reviewing the subject perturbations, three variations per subject are selected (The subject perturbations used are presented in Table~\ref{table:subpert}). We then query the models three times for each fact, using prompts that differ only in the subject token chunks, while the rest of the prompt structure remains identical. Figure \ref{subjectconsistencyfig} shows the prompt agreement levels of the models upon different subject perturbations. (the detailed agreement percentages for each model are provided in {Table~\ref{table:bothagreemneta}). The results reveal significant variation in prompt agreement, with many LLMs exhibiting low consistency, indicating that they produce differing responses to perturbed versions of the same subject entity. A positive trend is observed in more recent models, suggesting improved consistency over time. Instruction-tuned models, in particular, show comparatively higher prompt agreement. Among the evaluated models, ChatGPT and GPT-4 achieve the highest levels of consistency. On the other hand, GPT-\{2,3\}, T5 and OLMo 1B present low consistency, with OpenELM 270M demonstrating the lowest prompt agreement. 

\textbf{Property Perturbation.} We further evaluate the prompt agreement level of the models when queried with three different property lexicalizations. Different from the previous experiment (Subject Perturbation), the prompts used in this experiment consist of identical subject token chunks and differ only in the rest of the prompt, the \textit{property} to query (The prompts used are explained in subsection \ref{promptstrategysection}). Figure \ref{propertyconsistencyfig} presents the prompt agreement level upon different property perturbations (detailed agreement percentages for each model are provided in Table~\ref{table:bothagreemneta}). Similar to the previous experiment, the models demonstrate significant variation in prompt agreement, with more recent models demonstrating higher consistency in their responses. Furthermore, the higher consistency of instruction-tuned models is more evident in this experiment. Among the top performers, ChatGPT and GPT-4 show the highest consistency, followed by Mistral$_{I.}$, Mixtral$_{I.}$, and Llama-3$_{I.}$. On the other hand, GPT-\{2,3\} and OpenELM 270M demonstrate the lowest agreement levels. 

\subsection{Results} 

The findings from RQ1 (Assessment) can be summarized as: a) a persistent presence of outdated and incorrect responses across all models; and b) high sensitivity in the response generation process to minor perturbations in prompts, resulting in inconsistent outputs. These insights highlight the need for more robust training methodologies and updating mechanisms to enhance the internal knowledge representation of these models and improve their reliability.

\section{Enhancing LLMs' Reliability (RQ2)}

In this research question, we explore several approaches aimed at enhancing LLMs' performance as repositories of factual knowledge, including knowledge editing, Retrieval-Augmented Generation (RAG), and our proposed approach ENtity-Aware Fine-tuning (ENAF). We aim to address the shortcomings identified in RQ1 consisting of outdated knowledge, output inconsistency, and irrelevant outputs, providing a pathway to more reliable models. We select five LLMs to improve their performance: GPT-\{2,J\} due to generating a high percentage of outdated responses among models (42\% and 46\%, respectively); Llama-2$_{C.}$ and Mistral$_{I.}$ since, despite being relatively new models, provide outdated information to around 30\% of the questions; and OLMo 1B, a fully disclosed model and pre-training data, demonstrating high percentage of outdated responses (40\%) and low levels of output consistency (28\% over subject perturbations and 20\% over property perturbations).

\begin{table*}[h!]
\caption{Performance of SOTA methods (Knowledge Editing and RAG) for updating real-world outdated facts in five LLMs measured by the harmonic mean of efficacy success and paraphrase success.}
\label{table:edits_final}
\centering
\begin{tabular*}{38.9pc}{p{65pt} 
                       |p{35pt}<{\raggedleft} 
                       |p{25pt}<{\raggedleft} 
                       |p{25pt}<{\raggedleft}
                       |p{25pt}<{\raggedleft}
                       |p{25pt}<{\raggedleft}
                       |p{35pt}<{\raggedleft}
                       |p{35pt}<{\raggedleft}
                       |p{20pt}<{\raggedleft}
                       |p{20pt}<{\raggedleft}
                       |p{20pt}<{\raggedleft}}
\hline
\multirow{4}{*}{\textbf{(Year) Model}}& \multirow{3}{*}{\textbf{{\#Outdated}}} & \multicolumn{4}{c|}{\textbf{Knowledge Editing}} &\multicolumn{2}{c|}{\textbf{RAG}} & \multicolumn{3}{c}{\textbf{Vanilla Finetuning }} \\

  \cline{3-11} 
  &\multirow{3}{*}{\textbf{Facts}}& \multicolumn{2}{c|}{\textbf{Modifying}} & \multicolumn{2}{c|}{\textbf{Preserving}} &\multirow{2}{*}{\textbf{Retrieved}}&\multirow{2}{*}{\textbf{Gold}} & \multirow{2}{*}{\textbf{One}} & \multirow{2}{*}{\textbf{Two}} & \multirow{2}{*}{\textbf{Three}}\\
  
    && \multicolumn{2}{c|}{\textbf{Parameters}} & \multicolumn{2}{c|}{\textbf{Parameters}} &&&&\\

    \cline{3-6} \\[-8pt]
   &  & {ROME} & {MEMIT} & {SERAC} & {IKE} &\multirow{1}{*}{\textbf{Document}}&\multirow{1}{*}{\textbf{Document}}& \textbf{Epoch} &\textbf{Epoch} &\textbf{Epoch}\\
  \hline\\[-8pt]
  {\small(2019)} GPT-2 & 54 & 17\% & 33\% & 4\% & 49\% & 6\% & 17\% & 6\% & 5\% & 6\% \\
  {\small(2021)} GPT-J & 60 & 11\% & 83\% & 0\% & \underline{97\%} & 18\% & 29\% & 8\% & 10\% & 12\% \\
  {\small(2023)} Llama-2$_{C.}$ & 48 & 4\% & 77\% & 36\% & 18\% & 43\% & \underline{88\%} & 23\% &  17\% & 22\% \\
  {\small(2023)} Mistral$_{I.}$ & 41 & 0\% & 0\% & --- & \underline{92\%} & 84\% & \underline{99\%} & 21\% & 25\% & 8\%\\
  {\small(2024)} OLMo & 52 & --- & --- & --- & \underline{96\%} & 20\% & 43\% & 10\% & 10\% & 15\%\\
 \hline
 \multicolumn{11}{p{38pc}}{The "Knowledge Editing" shows the effectiveness of different methods in updating facts within the models. "RAG" columns report the performance when external information is incorporated at inference time for retrieved and gold-standard documents. The “Vanilla Finetuning” columns correspond to LoRA-based continued pretraining on the RAG documents for one, two, and three epochs. Underlined scores indicate the successful alignment of more than 85\% outdated knowledge.}\\
\end{tabular*}
\end{table*}

\subsection{State-of-the-Art Methods}

Several methods have been proposed to improve LLMs' performance in generating factual responses. Among these, knowledge editing and Retrieval-Augmented Generation (RAG) are two approaches that attempt to improve model accuracy. However, there are no comparative studies on these two categories of approaches to address the time-sensitive factual knowledge in LLMs \cite{zhang-etal-2023-large}. 

\textbf{Background.} Knowledge editing techniques involve modifying a fact in the model without requiring full retraining. These approaches aim for targeted adjustments to the model's knowledge. However, studies on editing techniques primarily rely on edit-target datasets of synthetically generated counterfacts, leaving a gap in understanding their effectiveness with real-world data across diverse domains \cite{zhang-etal-2023-large}. These methods also struggle with maintaining stability in the model's broader knowledge base, often leading to unintended side effects, such as introducing new errors or causing performance degradation on unrelated tasks \cite{pinter-elhadad-2023-emptying}. Furthermore, research in knowledge editing for LLMs has been narrowly focused on the updating operation  \cite{yao2023editing, mazzia2023survey}, often overlooking the importance of deletion and addition \cite{DIGNUM1992293}. RAG, on the other hand, integrates external knowledge sources, such as Wikipedia passages, into the model’s inference process \cite{gao2023retrieval,alghisi-etal-2024-fine-tune}. By retrieving up-to-date information dynamically, RAG aims to bypass the limitations of static pre-trained knowledge. While this approach shows promise in mitigating issues of outdatedness, it introduces its challenges. The reliance on external retrieval systems may lead to inefficiencies or inaccuracies, especially when high-quality retrieval data is unavailable \cite{cuconasu2024power}. Furthermore, RAG does not inherently address the deeper issue of outdated knowledge within the model itself, often leading to inconsistencies and conflicts between the retrieved and pre-trained information \cite{wu2024faithful}.

\textbf{Methods.} Regarding knowledge editing methods, we focus on two categories: approaches that modify the internal parameters of the model and those that preserve the original parameters by storing edits externally. Regarding parameter-based methods, \textbf{ROME} \cite{meng2022locating} identifies the specific parameters in the feed-forward layers responsible for a given fact. It then inserts a new key-value pair by formulating the update as a least-squares optimization problem. \textbf{MEMIT} \cite{meng2022mass} extends ROME by enabling multiple simultaneous edits across different layers in one operation, enhancing scalability for larger batches of edits. For methods that preserve the model's internal parameters, \textbf{SERAC} \cite{pmlr-v162-mitchell22a} stores new facts in an external memory. A classifier determines whether the question prompt matches the stored facts; if so, the model generates a response based on the stored facts, otherwise, the original LLM is used. \textbf{IKE} \cite{Zheng2023CanWE}, another parameter-preserving method, uses in-context learning by constructing prompts that include the question, the new fact, and a set of related examples. However, this technique may be less viable in dynamic real-world scenarios where new facts are not readily accessible. Finally, we compare these knowledge editing methods with RAG, which conditions the models' generation on relevant documents retrieved from external sources like Wikipedia. Although RAG has the potential of using up-to-date non-parametric knowledge, its success depends heavily on the quality and relevance of the retrieved documents. We also investigate noisy RAG, following previous studies \cite{cuconasu2024power} which suggest that introducing noise in a controlled manner can sometimes improve the retrieval process. Throughout our evaluations, ROME applies edits sequentially, MEMIT performs batch edits, and SERAC utilizes an external memory for each model consisting of corresponding new facts. For RAG, Wikipedia passages containing answers to outdated questions are compiled into a single file. We then use the LangChain framework\footnote{The LangChain framework is accessible \href{https://www.langchain.com/}{here}.} to split, encode, store, and retrieve the relevant passages within the RAG architecture. Regarding noisy RAG, for each question (e.g., an athlete’s team), we randomly sample a document from an irrelevant domain (e.g., countries’ politicians) and present it alongside the gold document to the model. As an additional baseline, we include a LoRA-based finetuning approach, in which only the low-rank adaptation parameters are updated on the next-token prediction objective. This baseline differs fundamentally from both editing and retrieval as it adapts the model to the updated factual text by continuing pretraining on the same set of documents used in the RAG setting, without modifying the full set of pretrained weights or relying on external retrieval. We use a LoRA rank of 32 and scaling factor $\alpha$ of 64. Training is performed with the AdamW optimizer, using a learning rate 1e-4 and batch size 1. To control the degree of adaptation, we train for one, two, and three epochs (columns “One/Two/Three Epoch” in Table~\ref{table:edits_final}).

\textbf{Harmonic Mean.} We evaluate the methods using the harmonic mean of two metrics: efficacy success and paraphrase success, following the literature \cite{meng2022locating, meng2022mass}. Efficacy success refers to the proportion of correct responses generated for the original question prompts, capturing the \textit{accuracy} for the edited knowledge (corresponding to RQ1 A). Paraphrase success, on the other hand, evaluates the model's ability to correctly respond to perturbed versions of the prompts (i.e. property perturbations only), serving as a \textit{consistency} metric (aligned with RQ1 B).

\begin{table}[t!]
\caption{Qualitative error analysis of state-of-the-art knowledge editing and RAG methods across five representative LLMs, based on human evaluation of model responses.}
\label{table:QA}
\centering
\begin{tabular*}{20.8pc}{p{63pt} |
                                p{23pt}<{\raggedleft\hangindent5pt}
                                p{23pt}<{\raggedleft\hangindent5pt}
                                p{23pt}<{\raggedleft\hangindent5pt}
                                p{23pt}<{\raggedleft\hangindent5pt}
                                p{23pt}<{\raggedleft\hangindent5pt}}
\hline
\multirow{2}{*}{\textbf{(Year) Model}} & \multirow{2}{*}{\textbf{C}{orrect}} & \multicolumn{3}{c}{\textbf{Incorrect}} & \textbf{M}odel \\
 \cline{3-5} 
  && \textbf{O}{utdated} & \textbf{G}{eneric}& \textbf{H}{alluc.} & \textbf{C}{ollapse}\\
 \hline
 \vspace{0.1pt}
 {{\small(2019)} \textbf{GPT-2}} &&&&&\\
   \hspace{0.2cm} {Van. FT}&5\%&	58\%&	18\%&	17\%&	2\%\\
  \hspace{0.2cm} {ROME}&19\%&8\%&4\%&66\%&4\%\\
   \hspace{0.2cm} {MEMIT}&25\%&45\%&18\%&12\%&0\%\\
 \hspace{0.2cm} {SERAC}&3\%&	0\%&	11\%&	23\%&	63\%\\
  \hspace{0.2cm} {IKE}&38\%&	56\%&	0\%&	6\%&	0\%\\
 \hspace{0.2cm} {RAG}&&&&\\
   \hspace{0.4cm} {\textit{Retrieved Doc}}&7\%&	42\%&	14\%&	37\%&	0\%\\
     \hspace{0.4cm} {\textit{Gold Doc}}&17\%&	51\%&	21\%&	11\%&	0\%\\
  \hline 
   \vspace{0.1pt}
  {\small(2021)} \textbf{GPT-J} &&&&\\
    \hspace{0.2cm} {Van. FT}&11\%&	55\%&	17\%&	17\%&	0\%\\

  \hspace{0.2cm} {ROME}&11\%&	0\%&	0\%&	0\%&	89\%\\
   \hspace{0.2cm} {MEMIT}&76\%&	2\%&	13\%&	7\%&	2\%\\
 \hspace{0.2cm} {SERAC}&0\%&	0\%&	0\%&	0\%&	100\%\\
  \hspace{0.2cm} {IKE}&87\%&	13\%&	0\%&	0\%&	0\%\\
 \hspace{0.2cm} {RAG}&&&&\\
   \hspace{0.4cm} {\textit{Retrieved Doc}}&17\%&	39\%&	12\%&	32\%&	0\%\\
     \hspace{0.4cm} {\textit{Gold Doc}}&29\%&	41\%&	19\%&	11\%&	0\%\\
 \hline  \vspace{0.1pt}

{\small(2023)} \textbf{Llama-2$_{C.}$} &&&&&\\
  \hspace{0.2cm} {Van. FT}&23\%&	69\%&	1\%&	6\%&	0\%\\
  \hspace{0.2cm} {ROME}&3\%&	0\%&	0\%&	0\%&	97\%\\
   \hspace{0.2cm} {MEMIT}&74\%&	3\%&	0\%&	23\%&	0\%\\
 \hspace{0.2cm} {SERAC}&35\%&	1\%&	0\%&	0\%&	64\%\\
  \hspace{0.2cm} {IKE}&15\%&	38\%&	47\%&	0\%&	0\%\\
 \hspace{0.2cm} {RAG}&&&&\\
   \hspace{0.4cm} {\textit{Retrieved Doc}}&45\%&	7\%&	48\%&	0\%&	0\%\\
     \hspace{0.4cm} {\textit{Gold Doc}}&86\%&	12\%&	1\%&	1\%&	0\%\\
     \hline  \vspace{0.1pt}
 {\small(2023)} Mistral$_{I.}$ &&&&&\\
   \hspace{0.2cm} {Van. FT}&18\%&	59\%&	3\%&	20\%&	0\%\\

   \hspace{0.2cm} {ROME}&0\%&	0\%&	0\%&	0\%&	100\%\\
   \hspace{0.2cm} {MEMIT}&0\%&	0\%&	0\%&	0\%&	100\%\\
  \hspace{0.2cm} {IKE}&89\%&	2\%&	8\%&	1\%&	0\%\\
 \hspace{0.2cm} {RAG}&&&&\\
   \hspace{0.4cm} {\textit{Retrieved Doc}}&83\%&	4\%&	11\%&	2\%&	0\%\\
     \hspace{0.4cm} {\textit{Gold Doc}}&98\%&	2\%&	0\%&	0\%&	0\%\\
      \hline  \vspace{0.1pt}
 {\small(2024)} OLMo \small{1B} &&&&\\
   \hspace{0.2cm} {Van. FT}&19\%&	42\%&	14\%&	24\%&	0\%\\

  \hspace{0.2cm} {IKE}&94\%&	4\%&	0\%&	3\%&	0\%\\
 \hspace{0.2cm} {RAG}&&&&\\
   \hspace{0.4cm} {\textit{Retrieved Doc}}&24\%&	32\%&	22\%&	22\%&	0\%\\
     \hspace{0.4cm} {\textit{Gold Doc}}&43\%&	27\%&	21\%&	9\%&	0\%\\
           \hline

\multicolumn{6}{p{20.pc}}{"Correct" indicates successful update of previously outdated response. "Incorrect" responses are categorized as a) \textit{Outdated}, where the model still outputs pre-training knowledge; b) \textit{Generic}, non-informative answers avoiding the factual response; c) \textit{Hallucination}, factually incorrect responses due to entity or attribute mix-ups; and, d) \textit{Model Collapse}, loss of generation ability, e.g., outputting special tokens. Parameter-editing methods (ROME, MEMIT) occasionally trigger collapse due to aggressive weight modifications. IKE generally achieves high correctness but often produces generic answers in some models. RAG exhibits strong dependence on retriever quality, with noisy retrieval increasing hallucinations and generic responses, while gold-standard documents markedly improve reliability. The Vanilla FT rows correspond to LoRA-based continued pretraining on the gold documents, which tends to preserve outdated knowledge and yield high rates of outdated or generic responses. This taxonomy complements the aggregate scores in Table~\ref{table:edits_final} by highlighting the distinct failure modes associated with each updating strategy.}\\

\end{tabular*}
\end{table}

\textbf{Results.} The results of our evaluation, summarized in Table~\ref{table:edits_final}, indicate that the effectiveness of knowledge editing methods is highly model-dependent. SERAC cannot be applied to OLMo and Mistral$_I$ due to the lack of support for these architectures. Similarly, parameter-based editing methods are not supported for OLMo, preventing direct parameter modifications. Additionally, such methods are ineffective in editing Mistral$_I$, as they result in outputting only special tokens following parameter updates. Among the edited models, GPT-J emerges as the most promising candidate for updates, achieving high success rates with both MEMIT and IKE. ROME and SERAC exhibit poor performance across models and are outperformed by MEMIT and IKE, respectively. Although three models (GPT-J, Mistral$_I$, and OLMo) show more than 90\% success with IKE, the knowledge editing methods overall demonstrate substantial limitations. While they have exhibited impressive performance on synthetic counterfactual datasets in the literature, their effectiveness diminishes significantly when applied to real-world updates, revealing serious gaps in their practical applicability. The LoRA-based continued-pretraining baseline (Table~\ref{table:edits_final} achieves consistently low harmonic-mean scores across models, despite being trained directly on the gold updated documents. The results indicate that finetuning is insufficient to override pretraining-time knowledge, and increasing the number of epochs does not lead to an improved performance. Thus, LoRA fine-tuning provides limited improvement and does not constitute a competitive alternative for targeted factual updates. Further comparison with RAG reveals that it demonstrates competitive performance, with generally higher alignment of instruction-tuned models which are designed to leverage contextual information more effectively. Nevertheless, its effectiveness is highly dependent on the quality of the retrieved documents, as anticipated. Although using gold documents significantly improves performance, it still falls short of achieving full alignment of any model. Additionally, our experiments with adding noise to the retrieval process did not yield considerable improvements. We observed no significant change in performance, and the results remained inconclusive.

\textbf{Qualitative Analysis.} To complement the results reported in Table \ref{table:edits_final}, we conduct a human evaluation of model responses to understand the impact of state-of-the-art methods on LLM behavior. Two researchers independently reviewed each response following an adaptation of the human evaluation framework proposed by \cite{mousavi-etal-2022-evaluation}. Responses are annotated as \textit{Correct} when a previously outdated answer (detected in RQ1) is successfully updated with the most current fact. Meanwhile, the incorrect outputs are grouped into four categories: a) \textit{Outdated} responses occur when the model continues to return outdated knowledge, hinting towards conflicts between parametric knowledge learnt during training and the applied intervention; b) \textit{Generic} responses are those where the model avoids generating a concrete factual answer, producing vague statements (e.g., “the president is the head of government”); c) \textit{Hallucinations} arise when the model fabricates or confuses facts, such as misattributing clubs to athletes or roles to political figures; and d) \textit{Model Collapse} captures cases where the model loses its ability to generate coherent text, producing empty strings, repetitions, or special tokens, a failure mode most commonly observed after aggressive parameter-editing methods.

The results of this analysis are presented in Table \ref{table:QA}. While the Vanilla Finetuning baseline occasionally corrects a small fraction of outdated answers, it predominantly preserves pre-training knowledge, leading to high rates of outdated or generic responses (e.g., 58–69\% outdated on GPT-2 and Llama-2$_{C.}$). Hallucinations remain relatively low compared to ROME and noisy RAG, but the method consistently fails to override the model’s parametric knowledge. ROME and MEMIT show very different behaviors depending on the model scale and architecture. In GPT-2, ROME mostly triggers hallucinations (66\%) with minimal model collapse (4\%), whereas MEMIT achieves more correct updates (25\%) but leaves nearly half of the answers still outdated (45\%). On GPT-J, by contrast, ROME leads to severe collapse (89\%), while MEMIT is far more stable, with most errors concentrated in the generic category (13\%). On Llama-2$_{C.}$, the model collapses almost completely (97\%) after applying ROME, while MEMIT improves correctness (74\%) but introduces a substantial fraction of hallucinations. On Mistral$_{I.}$, both ROME and MEMIT lead to total collapse (100\%). IKE achieves relatively high correctness across models, peaking at 94\% on OLMo, but it suffers from generic answers for Llama-2$_{C.}$, suggesting that in-context updates, while accurate, can reduce informativeness when contextual examples are not sufficiently specific. RAG shows a strong dependency on retriever quality. With noisy retrieval, models frequently produce hallucinations or generic statements (e.g., GPT-2, GPT-J, and OLMo), while gold documents substantially improve reliability (e.g., 86\% correct on Llama-2$_{C.}$, 98\% on Mistral$_{I.}$). SERAC demonstrates marked instability across settings. In GPT-2 it achieves only minimal success, with the majority of collapsing cases (63\%). On GPT-J it fails entirely, collapsing in 100\% of cases, and on Llama-2$_{C.}$, SERAC still produces collapse in 64\% of responses, underscoring its limited reliability. Taken together, these findings highlight that while both editing and retrieval approaches can mitigate outdated knowledge, they also introduce distinctive failure modes that practitioners must weigh when selecting an updating strategy.

\subsection{ENtity-Aware Finetuning (ENAF)}

Our evaluation of the state-of-the-art methods reveals limitations in the performance of existing approaches to enhancing LLMs' reliability, despite the partial success of some methods in aligning the models with updated factual knowledge.

To address this gap, we propose \textit{ENtity-Aware Fine-tuning (ENAF)}, a soft neurosymbolic approach in which the data used to train/fine-tune the model is augmented with structured representations of entities. Via ENAF, we aim to enhance the model’s ability to consistently and accurately recall information associated with specific entities by integrating symbolic entity representations into the learning process. This approach is motivated by the observation that accuracy and consistency in LLM outputs rely on different requirements. Accuracy depends on the model having access to up-to-date information from current sources, allowing it to generate factually correct responses. Consistency, on the other hand, is tied to the model’s understanding of both the entity being queried and the question itself. Usual (pre-)training and fine-tuning techniques typically lack a symbolic and structured representation of entities, often resulting in fragmented knowledge within the model, where different variations of the same entity are inconsistently represented. This is where a neurosymbolic representation proves valuable. By embedding structured entity representations, the model can map different perturbations of the same entity (e.g., lexical variations of a name) back to a unique symbolic reference, improving its ability to maintain consistent outputs across varied prompts.

\textbf{Experimental Setup.} We experiment with improving the models' consistency when queried about athletes' teams (28 subjects as soccer players, basketball players, and F1 drivers). We select OLMo 1B\cite{groeneveld2024olmo} as a fully (weight and training data) open-source model, and use Dolma\cite{soldaini2024dolma}, the collection of snapshots used to train the OLMo model family. Considering the huge size of the dataset (6.4 TB), we focus on the 'Wikipedia, Wikibooks' partition of the dataset (16 GB) as the source for time-sensitive factual knowledge, and select relevant documents based on two criteria A) the presence of any of the subjects (athletes) in the document title; or B) the presence of the subject entity in the document body for at least five times. The final set of documents contains a total of 9,501 relevant documents. We then augment this data by incorporating structured representations of entities in terms of tags (labels). Inspired by prior works in the literature \cite{raman-etal-2022-transforming}, we tagged each entity in our dataset by enclosing it between start and end delimiters \texttt{<TAG>} and \texttt{</>}. Regarding the representations, we experimented with five different scenarios:

\begin{table*}
\caption{Performance of our proposed soft neurosymbolic approach, ENtity-Aware Finetuning (ENAF), to enhance LLMs' input-bound consistency when prompted with subject and property perturbations.}
\label{table:enaf_agreement}
\centering
\begin{tabular*}{0.84\textwidth}{p{0.4\textwidth} |p{0.08\textwidth}<{\raggedleft} p{0.08\textwidth} <{\raggedleft} | p{0.08\textwidth}<{\raggedleft} p{0.08\textwidth}<{\raggedleft}}
\hline
\multirow{3}{*}{\textbf{Approach}} & \multicolumn{2}{c|}{\makecell{\textit{\textbf{GPT-2}}}} & \multicolumn{2}{c}{\makecell{\textit{\textbf{OLMo 1B}}}} \\
& \makecell{\textbf{Subject}} & \makecell{\textbf{Property}} & \makecell{\textbf{Subject}} & \makecell{\textbf{Property}} \\
& \makecell{\textbf{Agrmt. (\%)}} & \makecell{\textbf{Agrmt. (\%)}} & \makecell{\textbf{Agrmt. (\%)}} & \makecell{\textbf{Agrmt. (\%)}} \\
    \hline\\ [-8pt]
    \textbf{Pre-Trained Model}& 28\% & 24\% & 13\% & 44\%\\
    \textbf{Vanilla Finetuning}                     & 12\% & 30\% & 11\% & 33\%\\ \hline\\ [-8pt]
    
    \textbf{ENtity-Aware Finetuning}                             &&&&\\
    \hspace{0.2cm} \textit{Named Entity Tags (Subject \& Attribute)}            &  0\% & \textbf{50\%} &  6\% & \underline{35\%}\\
    \hspace{0.2cm} \textit{Selected Named Entity Tags (Subject \& Attribute)}           & 18\% & 26\% &  6\% & \textbf{43\%}\\
    \hspace{0.2cm} \textit{Normalized Entity Representation (Subject \& Attribute)}  & \underline{29\%} &\underline{33\%} & \textbf{33\%} &  9\%\\
    \hspace{0.2cm} \textit{ID Tags (Subject)}                                   & \textbf{38\%} & 22\% &  8\% &  6\%\\
    \hspace{0.2cm} \textit{ID Tags (Subject) + Selected Named Entity Tags (Attribute)}  & \underline{31\%} & \underline{28\%} & \underline{22\%} &  0\%\\
\hline
\multicolumn{5}{p{35pc}}{The table presents the models' agreement scores (Agrmt.) in vanilla fine-tuning and pre-trained baselines, compared against various structured entity representations under ENAF. Each model’s agreement score reflects its consistency in responding to perturbed prompts by subject and property. The structured representations include a) Named Entity Tags (for all entities); b) Selected Named Entity Tags (for frequent entities); c) Normalized Entity Representation (for all entities); d) ID Tags (for our 28 subject entities only),; and, a joint strategy, e) ID Tags (for our subject entities) + Selected Named Entity Tags (for frequent entities). Tagged token chunks are enclosed with start and end delimiters in the format "\texttt{<TAG>} token chunk \texttt{</>}", where \texttt{TAG} refers to either the named entity category or a unique ID. Bold and underlined scores indicate the highest agreement percentages within each model and ENAF strategy.}\\
\end{tabular*}
\end{table*}

\begin{enumerate}[I.]
    \item \textbf{\textit{Named-Entity Tags}}: where the dataset is augmented with the categorical tags of the detected Named Entities (NE). To detect the NEs we used an off-the-shelf tagger\footnote{We used SpaCy NER tagger with \href{https://spacy.io/models/en}{\texttt{'en\_core\_web\_trf'}} model.} (927,520 tagged entities with the NE vocabulary of 221,741). 
    
    \item \textbf{\textit{Selected Named-Entity Tags}}: where only the top 1.5\% frequent entities are tagged with the categorical labels (485,573 tagged entities with the NE dictionary of 3,479). This scenario was motivated by observing that more than 98\% of the NEs happen less than 30 times in the whole dataset. 
    
    \item \textbf{\textit{Normalized Entity Representation}}: where different lexicalizations of the same entity ($e_j'$,$e_j''$,\dots) is substituted with the root form of the entity ($e$). To retrieve the root form of the entities we used GPT-4 (Note that in this experiment the entity is replaced with the root form and not tagged). The motivation behind this experiment was to measure the impact of different lexicalizations (i.e. perturbations) of the entity on the models' consistency. 

    \item \textbf{\textit{ID Tags}}: where subject entities from our predefined list of 28 athletes were tagged with an ID. For each subject, we create a unique tag by concatenating their first and last name (e.g., \texttt{<CristianoRonaldo>}), and label all lexical variations of the subject by this consistent identifier. Unlike \textit{Normalized Entity Representation} which replaces the perturbations with a root form and removes the possibility of variations, ID tagging retains multiple variations of the same entity while associating them with a unique identifier. This is closer to real-world contexts, where subjects are often referenced by various names and aliases. Furthermore, tagging entities only with general labels such as NE categories (in this case "PERS") overlooks distinctions among different individuals (e.g., both "Cristiano Ronaldo" and "Lionel Messi" would fall under "PERS" but represent unique subjects). This experiment is designed to help the model preserve these distinctions while it establishes a neurosymbolic representation of each subject, allowing it to map different perturbations to a single reference.

    \item \textbf{\textit{ID Tags + Selected Named-Entity Tags}}: where both our 28 predefined subject entities and a selection of non-subject entities are tagged. Subject entities (i.e. athletes) are tagged with unique identifiers, as explained in the previous scenario; and non-subject entities are tagged with categorical NE labels if they are among the top 1.5\% in frequency. This hybrid tagging setup aims to help the model learn to distinguish specific subject entities while still recognizing and categorizing frequently occurring non-subject entities.

\end{enumerate}

Intuitively, the structured representations in ENAF resemble a star graph structure \cite{stargraph}, where different aliases of a subject are conceptually connected to a root node representing the unique identifier (e.g., \texttt{<CristianoRonaldo>}), or entities within a NE category are linked to a single node representing the category itself (e.g., \texttt{<PERS>}). In this setup, each alias or entity represents a leaf node, while the root node serves as the central hub. ENAF investigates whether symbolic and structured information can enhance model performance without requiring changes to the underlying architecture or training process of the models. While this work focuses on neurosymbolic tagging as a lightweight and efficient approach, future research may explore more advanced methods, such as explicit graph-based representations or deep symbolic learning, to capture richer relationships among entities and aliases.

To evaluate ENAF's effectiveness, we experiment with GPT-2 XL and OLMo 1B for each of the five scenarios. Experimenting with OLMo is motivated by the possibility of assessing the impact of fine-tuning with the structured version of the pre-training data instead of common unstructured text. Meanwhile, GPT-2 serves as a valuable baseline, as it has been used in evaluations of knowledge editing techniques, providing a reliable benchmark for assessing ENAF's impact in comparison to state-of-the-art methods. We further compare the observed results with standard vanilla fine-tuning with the unstructured version of the data used for ENAF. This comparison ensures that any observed changes in performance or behavior are due to the augmentations by ENAF rather than the models’ re-exposure to pre-training data. In all experiments, we apply the same fine-tuning setup for two epochs with LoRA using the AdamW optimizer (learning rate = 1e-4). No special modifications (e.g., layer freezing or custom losses) are introduced. The only difference across scenarios lies in the input formatting, where ENAF replaces plain text with structured entity representations. Appendix Table~\ref{table:enaf_examples} provides an example illustrating how the same factual statement is expressed under vanilla fine-tuning and ENAF configurations.

\textbf{Results.} Table~\ref{table:enaf_agreement} presents the results of ENtity-Aware Fine-tuning (ENAF) on model consistency upon subject and property perturbations for GPT-2 and OLMo in each scenario. We observe that generally, vanilla fine-tuning does not improve models' consistency. For instance, GPT-2’s subject agreement drops from 28\% (pre-trained model) to 12\%, while OLMo’s property agreement drops from 44\% (pre-trained model) to 33\%, suggesting that mere re-exposure to data does not inherently enhance input-bound consistency.  Augmenting the data with \textit{Named-Entity Tags} achieves the highest property agreement (50\%) for GPT-2 but performs poorly for both models in terms of subject agreement. This outcome aligns with the notion that generic tagging can help with property-related consistency but lacks the specificity needed for distinct subject identification. \textit{Selected Named Entity Tags} slightly improves OLMo’s property agreement (43\%), indicating that selective tagging of frequent entities enhances consistency over untagged or sparsely tagged data, though to a limited extent. On the other hand, \textit{Normalized Entity Representation} shows significant gains in subject agreement for OLMo (33\%) and maintains balanced subject and property agreement for GPT-2 (29\% and 33\%, respectively), suggesting that standardizing entity names enhances robustness against varied lexicalizations at inference time. Adding \textit{ID Tags} for subject entities outperforms all other scenarios in GPT-2’s subject agreement (38\%) and highlights the effectiveness of unique identifiers in creating stronger symbolic links for frequently referenced entities. However, this method shows limited improvements in property agreement, implying it primarily benefits subject-specific consistency. Meanwhile, \textit{ID Tags + Selected Named Entity Tags} demonstrates a balanced performance for subject agreement in OLMo, and both subject and property agreement in GPT-2, indicating it may serve as an optimal compromise. By using unique IDs for core subjects and tagging frequent non-subject entities, this setup helps the model distinguish specific subjects while acknowledging frequently occurring non-subject entities. Overall, these experiments show the potential of structured symbolic tagging of the data in promoting model consistency and suggest that entity-specific symbolic tagging allows LLMs to retain and recall consistent information more effectively than training/fine-tuning over bland textual data. This highlights the potential of symbolic and structured representations to capture and leverage entity relationships and semantic structures and enhance the consistency and reasoning capabilities of LLMs.

\begin{figure*}
    \centering
    \includegraphics[width=\textwidth,trim={1cm 0 0 0},clip]{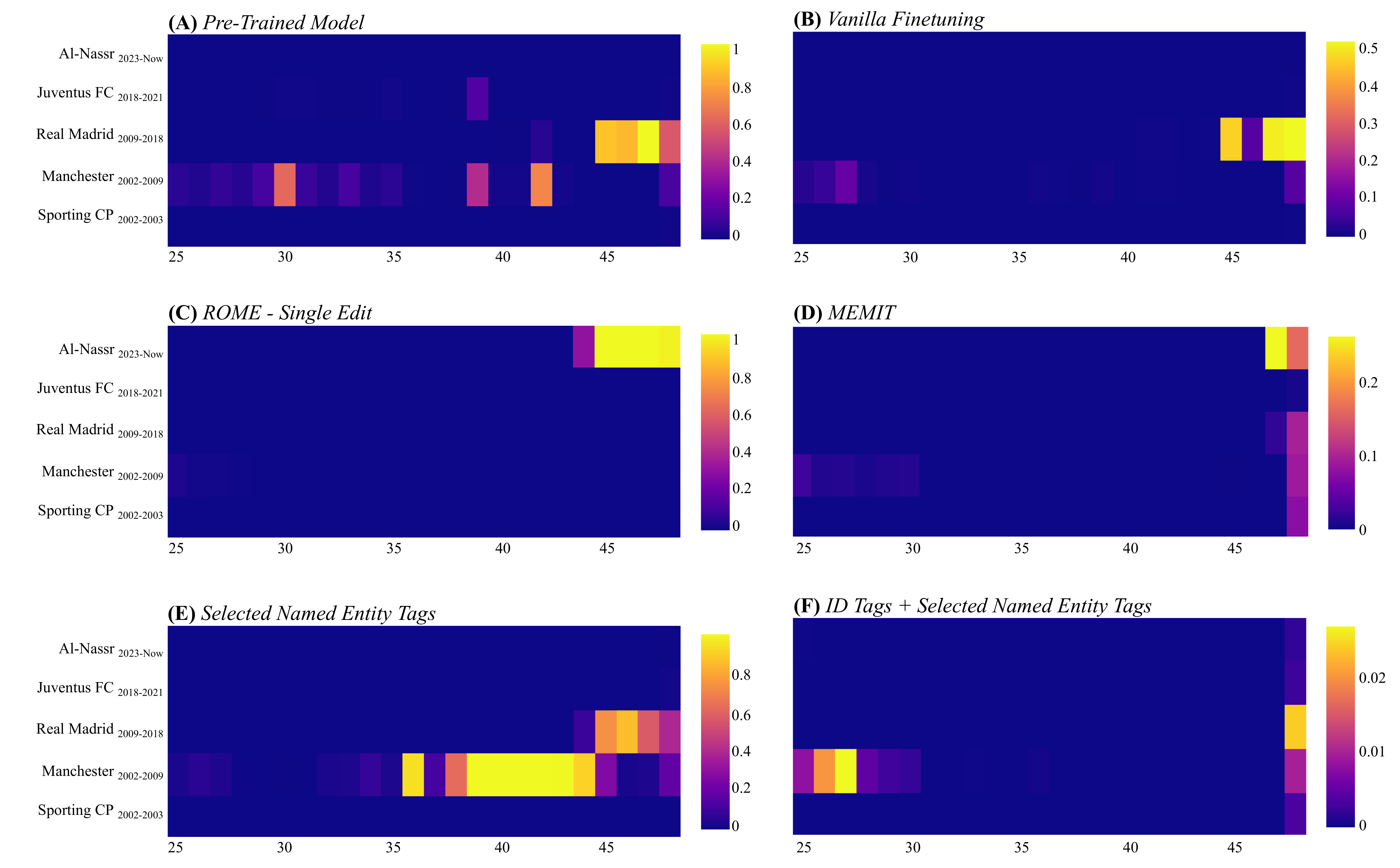}\\
    \caption{The layer-wise distribution of recall contributions for the subject entity \textit{Cristiano Ronaldo} associated with relevant attributes (i.e. affiliated clubs) across different fine-tuning and knowledge-editing strategies in GPT-2. "\textit{(A) Pre-Trained Model}" shows the contributions scattered across mid-to-upper layers. "\textit{(B) Vanilla Fine-Tuning}" skews the distribution towards a single attribute. Knowledge editing algorithms "\textit{(C) ROME}" and "\textit{(D) MEMIT}" concentrate the recall in the top layers for an edited association. ENtity-Aware Fine-tuning with "\textit{(E) Selected Named Entity Tags}" distributes recall contributions more evenly across layers, enhancing factual consistency. While "\textit{(F) ID Tags + Selected Named Entity Tags}" shifts contributions to the lower layers (responsible for factual retrieval for subject entity) and higher level, minimizing the impact of mid-layer contributions.}
    \label{fig:recall}
\end{figure*}

\textbf{Recall Process Analysis.} We investigate the impact of the enhancement methods on LLMs' layer-wise behavior when responding to factual questions. A recent study on the mechanisms of factual recall in LLMs \cite{recall} indicates that facts are primarily stored within feed-forward layers as subject-attribute associations, while attention mechanisms collaborate to refine and select the appropriate attribute to output. Inspired by \cite{recall}, we seek to explain how each method improves factual consistency across layers and support reliable responses across varied input prompts. Figures \ref{fig:recall} and \ref{fig:recall-lebron} present the recall process analysis for two subject entities, Cristiano Ronaldo, and LeBron James respectively. The heatmaps depict the layer-wise recall patterns across different strategies for each subject and the associated attributes (e.g., affiliated clubs). The heatmaps indicate that pre-trained models demonstrate scattered contributions with prominent attribute values in different layers. Meanwhile, vanilla fine-tuning results in a skewed distribution towards the most frequent attributes in the dataset. This behavior is also observed with ROME and MEMIT algorithms which strongly reinforce a targeted association. While these methods narrow the model’s recall for singular precise associations, they lack flexibility for scenarios where multiple associations must be maintained simultaneously (note that ROME and MEMIT aim at editing the model with the most recent information while the other methods are applied on OLMo pre-training data consisting of outdated/inconsistent information).

A common pattern observed in vanilla fine-tuning, ROME, MEMIT is the concentration of the models' recall almost exclusively to the last layers. In contrast, ENAF distribute the model’s recall contribution across multiple layers. \textit{Selected Named Entity Tags} spreads the recall contribution across mid and upper layers, preserving a more layered and distributed representation of factual knowledge within the model. Notably, \textit{ID Tags + Selected Named Entity Tags}, a hybrid approach with unique ID tags for subjects and NE tags for high-frequency non-subject entities, shifts the model’s recall contribution to the lower and upper layers. According to previous research \cite{recall}, lower layers play a crucial role in factual recall by retrieving all observed associations for the subject entity, while attention mechanisms in the upper layers refine and select the correct attribute value. Therefore, the increased contribution of the early layers when using subject ID tagging suggests that introducing IDs activates the association retrieval in early layers, i.e. the model’s recall of factual information for the subject entity. 



Overall, this analysis demonstrates that vanilla fine-tuning and editing methods tend to skew the distribution and concentrate recall contributions predominantly in the very last layers. In contrast, ENAF distributes the model’s contributions across multiple layers, preserving an equitable representation among attribute values. This finding highlights a trade-off between specificity and generalization: when using reliable, up-to-date data for a specific target task, methods that prioritize specificity by focusing on a desired value in the upper layers can offer a robust solution. However, when working with error-prone or problematic data such as the pre-training data often containing outdated or noisy information, maintaining the model’s generalizability becomes advantageous. By spreading recall across layers and allowing a broader representation of associations, ENAF helps retain flexibility and accuracy, ensuring the model remains resilient to variations in input.

\section{Conclusion}
We studied the reliability of LLMs as repositories of factual knowledge from two key perspectives: assessing their current performance (RQ1: Assessment) and exploring methods to enhance it (RQ2: Improvement). Our findings in RQ1 reveal significant areas needing improvement, as even the most recent models often produce outdated or incorrect responses. Additionally, LLMs display a pronounced sensitivity to slight perturbation in question phrasing, which can lead to inconsistent or irrelevant outputs. Our findings in RQ2 underscore the limitations of conventional fine-tuning and knowledge editing methods, which, while effective for isolated knowledge updates, often result in limited scalability and inconsistency across varied prompts. Our qualitative error analysis further reveals that these methods introduce distinct failure modes, ranging from generic answers to hallucinations and even full model collapse. These issues underscore the need for more robust training methodologies and dynamic updating mechanisms to enable LLMs to function as reliable repositories of knowledge.

To address these limitations, we introduced ENtity-Aware Fine-tuning (ENAF), a soft neurosymbolic approach that integrates structured representation of entities to improve model performance. Our work presents a trade-off between specificity-focused methods, suited for tasks with up-to-date and precise data, and generalizable approaches like ENAF, which offer greater resilience when faced with incomplete or outdated data. The DyKnow framework further complements this by enabling dynamic benchmarking, allowing us to assess the reliability of LLMs over time.

In conclusion, our results highlight the distinctive nature of LLMs compared to traditional knowledge repositories, emphasizing the importance of investigating the types of knowledge LLMs can manage effectively and the operations they support for querying and alignment. By advancing methods for structured fine-tuning and dynamic benchmarking, our work contributes to the ongoing effort to create more reliable and trustworthy language models for a range of knowledge-intensive applications.


\newpage

\appendix

\begin{figure}[h!]
    \centering
    \includegraphics[width=\columnwidth,trim={2.5cm 1cm 1cm 1cm},clip]{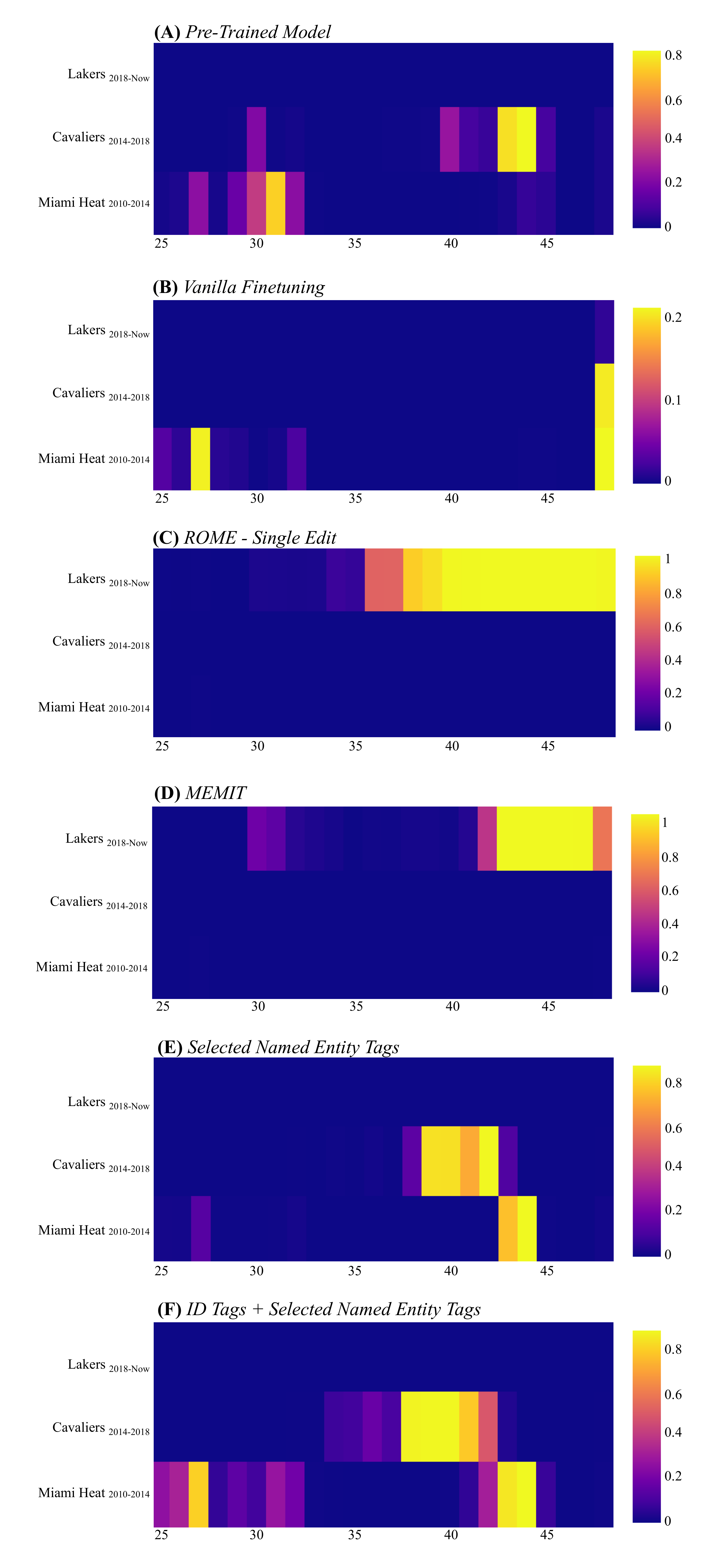}\\
    \caption{The layer-wise distribution of recall contributions for subject entity \textit{LeBron James} associated with affiliated clubs in GPT-2. "\textit{(A) Pre-Trained Model}" shows scattered contributions across layers. While "\textit{(F) ID Tags + Selected Named Entity Tags}" shifts the contributions to the lower layers (responsible for factual retrieval for subject entity) and higher levels, minimizing the impact of mid-layer contributions.}
    \label{fig:recall-lebron}
\end{figure}

\begin{table*}
\caption{This table presents aliases, alternative names, and variations for subject entities used in the "subject perturbation" analysis to assess the robustness of Large Language Models (LLMs) to variations in prompt phrasing. .}
\centering
\includegraphics[trim={0.2cm 6cm 0cm 0cm},clip,width=0.9\textwidth]{emnlpImages/tableofsubjects.pdf}
 \label{table:subpert}
\begin{tabular*}{38pc}{p{10pt}}

  \multicolumn{1}{p{37pc}}{}\\
 \end{tabular*}
 \end{table*}

\begin{table*}[h!]
\caption{Output consistency across LLMs when prompted with \textit{SUBJECT} and \textit{PROPERTY}  perturbations}
\label{table:bothagreemneta}
\centering
\begin{tabular*}{0.86\columnwidth}{p{90pt} |p{0.18\columnwidth}<{\raggedleft}
                                p{0.18\columnwidth}<{\raggedleft}}
\hline
\multirow{2}{*}{\textbf{(Year) Model}} & \textbf{Subject} & \textbf{Property} \\

 & \textbf{Agrmt. (\%)} & \textbf{Agrmt. (\%)} \\

    \hline\\ [-8pt]
    {\small(2019)} GPT-2                & 30\% &11\%\\
    {\small(2020)} GPT-3                & 10\% &9\% \\
    {\small(2020)} T5                   & 10\% &19\%\\
    {\small(2021)} GPT-J                & 31\% &25\%\\
    {\small(2022)} Bloom                & 41\% &25\%\\
    {\small(2022)} Flan-T5              & 48\% &49\%\\
    {\small(2023)} Llama-2              & 39\% &24\%\\
    {\small(2023)} Falcon               & 46\% &31\%\\
    {\small(2023)} Mistral              & 57\% &34\%\\
    {\small(2023)} Mixtral              & 50\% &29\%\\
    {\small(2024)} OLMo \small{1B}      & 28\% &20\%\\
    {\small(2024)} OLMo \small{7B}      & 39\% &23\%\\
    {\small(2024)} Llama-3              & 32\% &25\%\\
    {\small(2024)} OpenELM \small{270M} & 6\%  &4\%\\
    {\small(2024)} OpenELM \small{1.1B} & 29\% &22\%\\
    {\small(2024)} OpenELM \small{3B}   & 47\% &27\%\\
    \hline\\ [-8pt]
    {\small(2022)} ChatGPT              & 89\% &98\%\\
    {\small(2023)} GPT-4                & 91\% &94\%\\
    {\small(2023)} Llama-2$_{C.}$       & 72\% &82\% \\
    {\small(2023)} Falcon$_{I.}$        & 56\% &66\%\\
    {\small(2023)} Vicuna               & 56\% &69\%\\
    {\small(2023)} Mistral$_{I.}$       & 76\% &87\%\\
    {\small(2023)} Mixtral$_{I.}$       & 67\% &88\%\\
    {\small(2024)} Llama-3$_{I.}$       & 76\% &84\%\\
\hline
\multicolumn{3}{p{17pc}}{The table compares models' agreement scores (Agrmt.), which represent the percentage of consistent responses across perturbations in the subject or property within the prompt. LLMs below the line were prompted with an additional prefix "Answer with the name only". Instruction-tuned models demonstrate a comparatively higher prompt agreement across both perturbation types. Subscripts ${I.}$ and ${C.}$ stand for \textit{Instruct} and \textit{Chat}, respectively.}\\
\end{tabular*}
\end{table*}

\begin{table*}[h!]
\caption{Examples of entity representations used in vanilla fine-tuning and ENtity-Aware Fine-tuning (ENAF).}
\label{table:enaf_examples}
\centering
\begin{tabular*}{0.73\textwidth}{l |l}
\hline
\textbf{Entity Representation} & \textbf{Input Text} \\
    \hline\\ [-8pt]
    \textbf{Vanilla Finetuning} & \textit{Cristiano Ronaldo plays for Al-Nassr.} \\
    & \textit{CR7 is currently playing for Al-Nassr.} \\
    \hline\\ [-8pt]
    \textbf{ENtity-Aware Finetuning}                             &\\
     \multirow{2}{*}{\textit{Named Entity Tags}} & \textbf{\texttt{<PERS>}}\textit{Cristiano Ronaldo}\textbf{\texttt{</>}} \textit{plays for} \textbf{\texttt{<ORG>}}\textit{Al-Nassr}\textbf{\texttt{</>}}.\\
     & \textbf{\texttt{<PERS>}}\textit{CR7}\textbf{\texttt{</>}} \textit{is currently playing for} \textbf{\texttt{<ORG>}}\textit{Al-Nassr}\textbf{\texttt{</>}}. \\
     \\
    \textit{Normalized Entity} & \textit{Cristiano Ronaldo plays for Al-Nassr.}\\
     \textit{Representation}& \textit{\textbf{Cristiano Ronaldo} is currently playing for Al-Nassr.} \\
     \\
    \multirow{2}{*}{\textit{ID Tags}}
     & \textbf{\texttt{<CristianoRonaldo>}}\textit{Cristiano Ronaldo}\textbf{\texttt{</>}} \textit{plays for Al-Nassr.}\\
     & \textbf{\texttt{<CristianoRonaldo>}}\textit{CR7}\textbf{\texttt{</>}} \textit{is currently playing for Al-Nassr.} \\
     \\
    \textit{ID Tags + Selected}
     & \textbf{\texttt{<CristianoRonaldo>}}\textit{Cristiano Ronaldo}\textbf{\texttt{</>}} \textit{plays for \textbf{\texttt{<ORG>}}\textit{Al-Nassr}\textbf{\texttt{</>}}.}\\
    \textit{Named Entity Tags}& \textbf{\texttt{<CristianoRonaldo>}}\textit{CR7}\textbf{\texttt{</>}} \textit{is currently playing for \textbf{\texttt{<ORG>}}\textit{Al-Nassr}\textbf{\texttt{</>}}.} \\
\hline
\multicolumn{2}{p{30.5pc}}{Examples of different entity representations for the same factual statement (‘Cristiano Ronaldo plays for Al-Nassr’) across training configurations. Vanilla fine-tuning uses unstructured text, while ENAF introduces structured representations: \textit{Named Entity Tags} mark entity categories; \textit{Normalized Entity Representation} unifies lexical variants into a canonical form; \textit{ID Tags} link mentions to unique identifiers; and, the hybrid strategy \textit{ID Tags + Selected NE Tags} combines both. These structured inputs serve as soft neurosymbolic anchors, helping the model align different lexicalizations (e.g., ‘Cristiano Ronaldo’ vs. ‘CR7’) and improve consistency in factual responses.}\\
\end{tabular*}
\end{table*}

 




\vfill

\end{document}